\documentclass{bmvc2k}

\usepackage[T1]{fontenc}
\usepackage{graphicx}

\usepackage{amsmath}
\usepackage{amssymb}
\usepackage{lipsum}
\usepackage{marginnote} 
\usepackage{booktabs}
\usepackage{cleveref}
\usepackage{caption}

\def\NN{ \mathbb{N} }
\def\SA{ \mathbb{S} }
\def\TT{ \mathbb{T} }
\def\RR{ \mathbb{R} }

\def\SO3{\mathcal{SO}(3)}

\title{Are Euler angles a useful rotation parameterisation for pose estimation with Normalizing Flows?}

\def\etal{\emph{et al}\bmvaOneDot}

\addauthor{Giorgos Sfikas}{gsfikas@uniwa.gr}{1}
\addauthor{Konstantina Nikolaidou}{konstantina.nikolaidou@ltu.se}{2}
\addauthor{Foteini Papadopoulou}{fpapadopoulou@uniwa.gr}{1}
\addauthor{George Retsinas}{gretsinas@central.ntua.gr}{3}
\addauthor{Anastasios L. Kesidis}{akesidis@uniwa.gr}{1}

\addinstitution{
 Dpt. of Surveying and Geoinformatics Engineering,
 University of West Attica,\\
 Athens, Greece
}
\addinstitution{
 Dpt. of Computer Science,
 Luleå University of Technology,\\
 Luleå, Sweden
}
\addinstitution{
 School of Electrical and Computer Engineering,
 National Technical University of Athens,\\
 Athens, Greece
}

\runninghead{Sfikas \etal}{Are Euler angles a useful rotation parameterisation?}

\begin{document}
\maketitle

\begin{abstract}

Object pose estimation is a task that is of central importance in 3D Computer Vision.
Given a target image and a canonical pose, a single point estimate may very often be sufficient; however, a probabilistic pose output is related to a number of benefits
when pose is not unambiguous due to sensor and projection constraints or inherent object 
symmetries.
With this paper, we explore the usefulness of using the well-known Euler angles parameterisation as a basis for a Normalizing Flows model for pose estimation.
Isomorphic to spatial rotation, 3D pose has been parameterized in a number of ways, either in or out of the context of parameter estimation.
We explore the idea that Euler angles, 
despite their shortcomings, may lead to useful models in a number of aspects,
compared to a model built on a more complex parameterisation.

\end{abstract}
\section{Introduction}

Estimating the pose of objects from a single RGB image is an important and challenging problem in computer vision~\cite{mahendran20173d,balntas2017pose,hoque2021comprehensive}.
It is closely connected to several scientific challenges, such as navigation in 3D scenes, augmented and virtual reality, robotic manipulation, and autonomous driving~\cite{hoque2023deep,tremblay2018deep}.
In its more general form, the problem of estimating object pose encompasses 6 degrees of freedom (DOF); 3DOF for spatial position and 3DOF for spatial rotation.
In this work, we focus on 3D object pose estimation, i.e., estimation of object spatial rotation~\cite{balntas2017pose}.
Furthermore, we are interested in obtaining a probability density function instead of just a single point estimate, as in standard rotation regression  (e.g.~\cite{liao2019spherical}).
The motivation for this choice is that a probabilistic formulation of the problem constitutes an elegant way to describe both ambiguities due to inherent uncertainty of pose given a single image related to the characteristics of central projection,
as well as uncertainty arising due to the geometry of the object itself.
A deterministic prediction will be inadequate given the multimodal nature of the problem, resulting in oversimplified solutions that ignore symmetry and uncertainty. 
Hence, in general there is an irreducible, aleatoric component of uncertainty that is incorrect to disregard.
Furthermore, the probabilistic approach subsumes the non-probabilistic case as the latter can always be described as a Dirac PDF around the point estimate.
Several parametric distributions have been proposed for probabilistic modeling of rotations, including the von Mises distribution for Euler angles ~\cite{liu2023delving, Prokudin_2018_ECCV}, the Bingham distribution for quaternions~\cite{deng2022deep, gilitschenski2019deep}, and the Matrix Fisher distribution for rotation matrices~\cite{mohlin2020probabilistic}. 
While these models are statistically grounded, their unimodal nature limits their ability to accurately capture symmetric objects, which are common in real-world environments~\cite{liu2023delving}.

In the proposed approach, we use Normalizing Flows (NFs) as our model for density estimation.
NFs are a powerful and flexible framework for learning complex distributions through invertible mappings from simple base distributions. 
Although NFs have been successfully applied in Euclidean spaces~\cite{falorsi2020neural}, their extension to the non-Euclidean manifold $\SO3$ remains underexplored. 
Part of existing approaches, such as ReLie~\cite{falorsi2019reparameterizing} and ProHMR~\cite{kolotouros2021probabilistic}, 
attempt to project latent Euclidean flows onto $\SO3$.
Other methods~\cite{falorsi2020neural, lou2020neural, mathieu2020riemannian} develop flows on general Riemannian manifolds without exploiting the geometric structure of $\SO3$, often resulting in limited representational capacity and suboptimal performance~\cite{liu2023delving}.
Liu \etal~\cite{liu2023delving} suggest constructing a flow as a composition of M\"obius transforms and the newly introduced quaternion affine layer.
Their M\"obius transform works as a modification of the original use for spheres in~\cite{rezende2020normalizing}, where a coupling approach uses one column of a rotation matrix as input for the conditioner, while the coupling transformer acts on a second column.
As the output of the transform is not necessarily a valid rotation matrix, in each flow step, transform parameters are projected so that the result stays in $\SO3$.
The quaternion affine layers act via a linear transform
over a $\RR^4$ representation of the rotation quaternion. 
Again, the result is projected to the rotation manifold.
\textit{In this paper, we explore the following idea:}
Given the complexity of the rotation matrix parameterisation in the context of estimation with NFs, can it be beneficial to employ a simpler parameterisation of rotation?
We hypothesize that a simpler parameterisation may constitute a simpler objective for our learning model.
Ideally, we would want a lower dimensionality parameterisation, and one that would not require per-flow step reprojections to the rotation manifold as in~\cite{liu2023delving}.
Euler angles are a parameterisation that fits this description~\cite{forstner2016photogrammetric}; they are the simplest parameterisation possible, using $3$ unconstrained angles of rotation about fixed axes, a number that matches the intrinsic degrees of freedom of spatial rotation.
Euler angles have been (rightfully) critiqued due to well-known shortcomings
(cf. Section \ref{sec:proposed}).
However, they are very much in use in specific contexts, for example, in Photogrammetry~\cite{stylianidis2020measurements}.
%
We test our assumption with experimental trials on a number of datasets that cover diverse object geometries that may encompass complex symmetries.
We show that the proposed model will lead to better density estimation compared to other models, including NFs that use rotation matrices for parameterisation~\cite{liu2023delving}.
We also check the effectiveness of our model on cases where data lie on subsets of $\SO3$ that are related to gimbal lock singularities for Euler angles.
Qualitative results suggest that Euler angles are related to different inductive biases that may often be suboptimal (cf. e.g. Figure~\ref{table:qualitative}).
However, overall, we obtain slightly better fits than rotation matrix parameterisation NFs.
Perhaps not unsurprisingly, an exception to this rule is datasets with ground truth heavily concentrated around points of Euler angle singularity.


The remainder of the paper is structured as follows.
Section~\ref{sec:flowcircular} discusses extensions of flow models to non-Euclidean domains, including $\SO3$.
The proposed alternative is presented in Section \ref{sec:proposed},
including a succinct introduction on Euler angles and their advantages and shortcomings as a parameterisation of rotation.
Experimental results are presented in Section \ref{sec:experiments},
and we close the paper with our conclusions and future work in Section \ref{sec:conclusion}.

\section{Normalizing Flows on circular and spherical domains}
\label{sec:flowcircular}

This section reviews using NFs on non-Euclidean domains like $\SA$ or $\SO3$.
Much of the framework has been laid down in \cite{rezende2020normalizing},
while \cite{liu2023delving} have adapted the toolset for use with the rotational manifold.

\paragraph{Elements.}

The basic idea in NFs is as follows.
Consider first a simple Gaussian model for density estimation:
    $p(x; \mu, \Sigma) \propto \exp\{-\frac{1}{2}||g_{\mu,\Sigma}(x)||^2\}$,
where $g_{\mu,\Sigma}(x) = \Sigma^{-1/2}(x-\mu)$.
In NFs, the affine $g_{\bullet}(x)$ is replaced by a non-linear transformation, 
and in particular, a special class of invertible neural networks.
Of course, a closed-form Maximum Likelihood (ML) solution is out of the question if $g_\theta(\cdot)$ 
is a neural network, but we can still use the arsenal of 
gradient-based optimisation to obtain reasonable estimates (much like in the rest of Deep Learning).
Furthermore, invertibility and differentiability of $g_\theta(\cdot)$ ensure that processes like sampling and evaluation with respect to the learned distribution are tractable and conceptually very straightforward.
For example, sampling amounts to simply drawing a sample from a standard Gaussian and transforming it according to the inverse transformation $g^{-1}_\theta(\cdot)$.
So, the likelihood model in NFs is:
    $p(x; \theta) \propto \exp\{-\frac{1}{2}||(g_{\theta}(x)||^2\}$,    
where $g_\theta(x)$ is defined as a non-linear bijection.
\footnote{The larger part of the Normalizing flow literature
follow the convention that the normalizing transform $g$ is 
the ``inverse'' direction and $g^{-1}$ is the ``forward'' direction~\cite[Ch.16]{prince2023understanding}.
The current convention is used to simplify formulae and follow the implementation of~\cite{liu2023delving}.
Hence, here $g$ represents the normalizing direction and $g^{-1}$ represents the generative direction~\cite[Ch.16.1.2]{prince2023understanding}
}
The major benefit of non-linearity over $g_\theta(x)$ is that we can model any distributional form (given constraints for the form of $g$)~\cite{kobyzev2020normalizing}.

\paragraph{Conditional flows.}

NFs can be used in a supervised setting, 
and similar to the paradigm of GANs or Diffusion models, here, too, 
we have a conditional counterpart of the basic model.
In terms of density estimation, we aim to learn $p(x|z)$ instead of $p(x)$.
In the current application context of pose estimation, condition $z$ is an input image (or image plus canonical pose), and the expected outcome is a distribution over likely poses.

Modeling of joint or more complex distributions follows similar considerations,
whereas $p(x_1,x_2)$ can be written as $p(x_2|x_1)p(x_1)$, and we need a way to model conditional PDFs with flows.
Coupling flows and autoregressive flows are two alternatives that may be used to model conditional flows.

\paragraph{Handling circular and spherical topologies.}
NFs on bounded and circular domains like circles ($\SA$) or Tori ($\TT$) require special considerations.
First, the Gaussian as a base distribution is inadequate, as it implies an unbounded, Euclidean topology, for any $n\in \NN$ and $\RR^n$.
Second, flow layers also need to be carefully designed
so that the circular character of these topologies is respected.
For example, transformations that are defined over $\SA$ (the circle) should take into account the periodicity of the domain.

Rezende \etal~\cite{rezende2020normalizing} have pioneered working with NFs on circles and tori, and have introduced a number of suitable flow layers to this end.
A number of options that satisfy the conditions set by \cite{rezende2020normalizing}
among which the M\"obius transformations~\cite{kato2015m}. 
Other options are circular splines and non-compact projections,
the latter of which is a complex form of the M\"obius transform
\cite[Appendix H]{rezende2020normalizing}.
The M\"obius transformation~\cite{kato2015m} is defined as follows.
Given a D-dimensional sphere $\SA^D$, we can consider it as the locus
of points with unitary distance from the origin in $\RR^{D+1}$.
Let $w \in \RR^{D+1}$ be a vector with $||w|| < 1$,
which will act as the parameter of the transformation.
Then a point $x \in \RR^{D+1}$ is mapped according to the M\"obius transform
with parameter $w$ as follows:
\begin{equation}
    g_w(x) = \frac{1-||w||^2}{||x-w||^2}(x-w) - w.
\end{equation}
It can be easily checked that $w = 0$ corresponds to an identity transformation.
The part of the sphere that is close to the parameter point $w$ is expanded as a result,
and the rest is contracted.

One crucial property that, in the context of flow expressivity, proves to be a drawback is that the group of M\"obius transformations is closed
under composition~\cite[Theorem 2]{kato2015m}.
There are two workarounds concerning this point: either use a convex combination of two or more M\"obius transforms, or compose a flow using other flow layers other than M\"obius transforms
in-between them.
Both solutions are used in practice.
Unlike the M\"obius transform itself, there is no closed-form for the inverse of the convex combination of M\"obius transforms.
However, it can be numerically inverted with precision $\epsilon$ using bisection search within $\mathcal{O}(log\epsilon^{-1})$ iterations.

\vspace{-.5cm}
\paragraph{Defining flows on $\SO3$.} 
In Liu \etal~\cite{liu2023delving}, a NF for $\SO3$ is defined.
This is defined as a cascade of M\"obius layers and Quaternion affine layers.
In order to capture the geometric constraints of the $\SO3$ manifold,
the M\"obius transforms are defined in a very specific manner,
acting in each flow layer over a single column of the rotation matrix 
$R = [r_1\phantom{k}r_2\phantom{k}r_3]$, where $r_i$ are the columns of $R$.
As the column to be transformed, say $r_2$, must adhere to being unit-norm,
it is treated as a point on the two-dimensional sphere $\SA^2$.
At the same time, $r_2$ must be orthogonal to $r_1$ and $r_3$.
The strategy of~\cite{liu2023delving} is to use one of the remaining columns as the conditioner, meaning that it will act as input to the coupling NN, 
which will output M\"obius transform parameters 
. 
The transformer column is ensured to be orthogonal to the conditioner column by applying a suitable projection over the M\"obius transforms parameters ($w = w' - r_1[r_1\cdot w'])$, where $w$ and $w'$ are output and input parameters, respectively.
The other column, say $r_3$, will be computed as the outer product of the rest ($r_3 = r_1 \times r_2)$.
With this, all rotation matrix constraints are met.
In each subsequent M\"obius layer, the role of each column, i.e., whether it is a conditioner, a transformer, or it is computed via the outer product, is changed in order to ensure that no single column has any special bias.
Quaternion affine layers act by defining a linear transformation over a real-valued representation of the rotation quaternion.
Judging by the presented results of~\cite{liu2023delving}, between M\"obius and Quaternion affine, M\"obius transforms arguably contribute the more ``added value'' to the model.


\section{Proposed Euler Angles-based Flow}
\label{sec:proposed}

\paragraph{Euler angles.}
NFs work by defining a transformation over data, the characteristics and motivation of which we have already described.
In the context of the current application, our data are elements of the $\SO3$ manifold.
Various ways have been put forward to parameterize $\SO3$, each one with its advantages and disadvantages.
In~\cref{sec:flowcircular} we saw that previous work uses the Rotation matrix~\cite{liu2023delving}.

Perhaps the simplest parameterisation available is by using ``Euler angles''.
Their motivation is related to a celebrated theorem by Euler, which states that any two independent orthonormal coordinate frames 
can be related by a sequence of rotations about coordinate axes, where no two successive rotations may be about the same axis~\cite[Section 4.3]{kuipers1999quaternions}.
Crucially, \emph{not more than three} successive rotations are necessary
to describe any arbitrary rotation.
Also, the axes of rotation may be chosen arbitrarily.
A straightforward choice of axes is the coordinate frame axes, $X,Y,Z$.
Hence, given a fixed sequence of axes, we only require three angles
to describe a rotation, which are called Euler angles~\cite{kuipers1999quaternions,forstner2016photogrammetric}.
A usual convention in Photogrammetry is to denote rotations about $X,Y,Z$
with letters $\omega, \phi, \kappa$ respectively (for example,~\cite{karras1999metric,stylianidis2020measurements}).
These define a set of rotation matrices:

\begin{equation}
    \begin{aligned}
        R_\omega &= \begin{bmatrix}
            1 & 0 & 0 \\
            0 & \cos\omega & \sin\omega \\
            0 & -\sin\omega & \cos\omega \\
        \end{bmatrix},
        R_\phi &= \begin{bmatrix}
            \cos\phi & 0 & -\sin\phi \\
            0 & 1 & 0 \\
            \sin\phi & 0 & \cos\phi \\
        \end{bmatrix},
        R_\kappa &= \begin{bmatrix}
            \cos\kappa & \sin\kappa & 0 \\
            -\sin\kappa & \cos\kappa & 0 \\
            0 & 0 & 1 \\
        \end{bmatrix}
    \end{aligned}
\end{equation}
and the full rotation is written as a composition of the three composing rotations.
Corresponding to the sequence of rotation around $X,Y$, then $Z$, we have
    $R = R_\kappa R_\phi R_\omega$.
An inverse relation, taking us from the rotation matrix to Euler angles, can be elaborated.
However, here manifests the problem of ``gimbal lock'': there exist areas that are not matched to a set of Euler angles in a unique manner.
For example, for $\phi=\frac{\pi}{2}$
we can only determine the difference of Euler angles $\omega-\kappa$, but not each one of them separately:
\begin{equation}
    R = \begin{bmatrix}
        0 & cos(\omega-\kappa) & sin(\omega-\kappa) \\
        0 & -sin(\omega-\kappa) & cos(\omega-\kappa) \\
        1 & 0 & 0 \\
    \end{bmatrix}
\end{equation}
A similar result holds for $\phi = -\frac{\pi}{2}$.
In our implementation, we redefine these relations to be bijective: Euler angles to rotation matrices are forced to be unambiguous definitions.
For example, when converting rotation matrices to Euler angles and $cos\phi = 0$, we use the convention $\kappa=0$
and the remaining degree of freedom is assigned effectively to $\omega$.
Note that the measure described here by no means ``solve'' gimbal lock in Euler angles.
Our point of view is that we may be able to take advantage of the positive aspect related to Euler angles (i.e., their simplicity to implement),
under the premise that the greater part of the $\SO3$ manifold does not suffer from this singularity.

\begin{table*}[h]
\centering
\caption{
Comparisons of the proposed Euler angles-based model (``Ours'') versus prior models.
The quality of fit to pose data is evaluated using a log-likelihood measure.}
\label{table:synthetic}
\begin{tabular}{l c c c c c}
\\ \toprule
log likelihood $\uparrow$ & avg. & peak & cone & cube/fisher24 & line \\
\midrule
Riemannian~\cite{mathieu2020riemannian} & 5.82 & 13.47 & 8.82 & 1.02 & -0.026 \\
ReLie~\cite{falorsi2019reparameterizing} & - & - & 5.32 & 3.27 & -6.97 \\
IPDF~\cite{murphy2021implicit} & 4.38 & 7.30 & 4.75 & 4.33 & 1.12 \\
Mixture MF~\cite{mohlin2020probabilistic} & 6.04 & 10.52 & 8.36 & 4.52 & 0.77 \\
Moser Flow~\cite{rozen2021moser} & 6.28 & 11.15 & 8.22 & 4.42 & 1.38 \\
M\"obius~\cite{liu2023delving} & 7.28 & 13.93 & 8.99 & 4.81 & 1.38 \\
M\"obius+Affine Flow~\cite{liu2023delving} & 7.28 & 13.93 & 8.99 & 4.81 & 1.38 \\
\hline
Euler angles flow (ours) & \textbf{8.07} & \textbf{15.09} & \textbf{9.35} & \textbf{5.71} & \textbf{2.14} \\

\bottomrule
\end{tabular}
\end{table*}

\paragraph{Defining flows with Euler angles.}
The proposed flow uses the M\"obius transform as its backbone.
Compared to a flow that uses rotation matrices (cf. sec.~\ref{sec:flowcircular}, the implementation is arguably much more straightforward.
In each flow step, our input is the three Euler angles $x = (\omega, \phi, \kappa)$.
The domain of each is taken to be a unit sphere $\SA^1$, so $\omega,\phi,\kappa$ all take values $\in [0, 2\pi]$.
Equivalently, we have $x \in \SA^1 \times \SA^1 \times \SA^1$.
Our strategy is to use a coupling flow (cf. Section \ref{sec:flowcircular}),
where in each flow step, one subset of the Euler angles will correspond to the conditioner, and the other corresponds to the transformer.
Assuming angle $\kappa$ is being transformed in a given step, we have:
\begin{align}
    \omega' = \omega \nonumber,
    \phi' = \phi \nonumber,
    \kappa' = \sum_{i=1}^K \rho_i [\frac{1-||w_i||^2}{||\kappa-w_i||^2}(\kappa-w_i) - w_i],
\end{align}
where $\sum_{i=1}^K\rho_i = 1$ and $\rho_i \geq 0$ for all $i\in[1,K]$.
As parameters of the transformer, we have:
    $\{\rho_i, w_i\}_{i=1}^K = \phi[cos\omega, sin\omega, cos\phi, sin\phi]$,
where $\phi[\cdot,\cdot,\cdot,\cdot]$ is defined as a standard (non-invertible) neural network, 
which will output parameters for the $K$ combined M\"obius kernels.
Akin to a minimal version of positional encoding~\cite{mildenhall2021nerf}, we have mapped conditioner Euler angles as $\theta \mapsto (cos\theta, sin\theta)$,
in order to allow the network $\phi[\cdot,\cdot,\cdot,\cdot]$ to use the fact that the domain of the angles is periodic.
The output parameters are fed to the coupling neural network.
In effect, the above equations describe a coupling transform where the transformer is defined via a convex combination of M\"obius transforms over $\SA^1$.
Of course, this formulation will allow transformation only for the Euler angle $\kappa$.
In order to allow transformations for the other two Euler angles, we apply permutations of the role of each angle, in a round-robin manner.

The end result is a flow model that aims to perform estimation in a distinctly simpler manner than the approach based on rotation matrix parameterisations (cf. Section \ref{sec:flowcircular},~\cite{liu2023delving}).
The dimensionality of the target quantity is lower, where we have a single angle in each step, and unconstrained, meaning that no projections are required.
We argue that this scheme is more convenient for learning and inference, both in the sense of a learning objective, as well as in terms of computational load.

\section{Experimental Results}
\label{sec:experiments}

We conduct experiments to evaluate our proposed method, deploying Euler angles on NFs.
We have run experiments on a collection of different datasets that cover different aspects of the task.
Datasets ``Synthetic'' and ``Gimbal'' are used to gauge performance on unconditional density estimation on rotation data.
Datasets ``SYMSOL I'' and ``ModelNet10-SO3'' are benchmarks that involve image input data, covering both objects that lend to inherently multimodal as well as unimodal pose ground truths.

\begin{figure}[ht]
    \centering
    \begin{minipage}{0.49\linewidth}
        \centering
        \begin{tabular}{ccc}
            \includegraphics[width=0.28\textwidth]{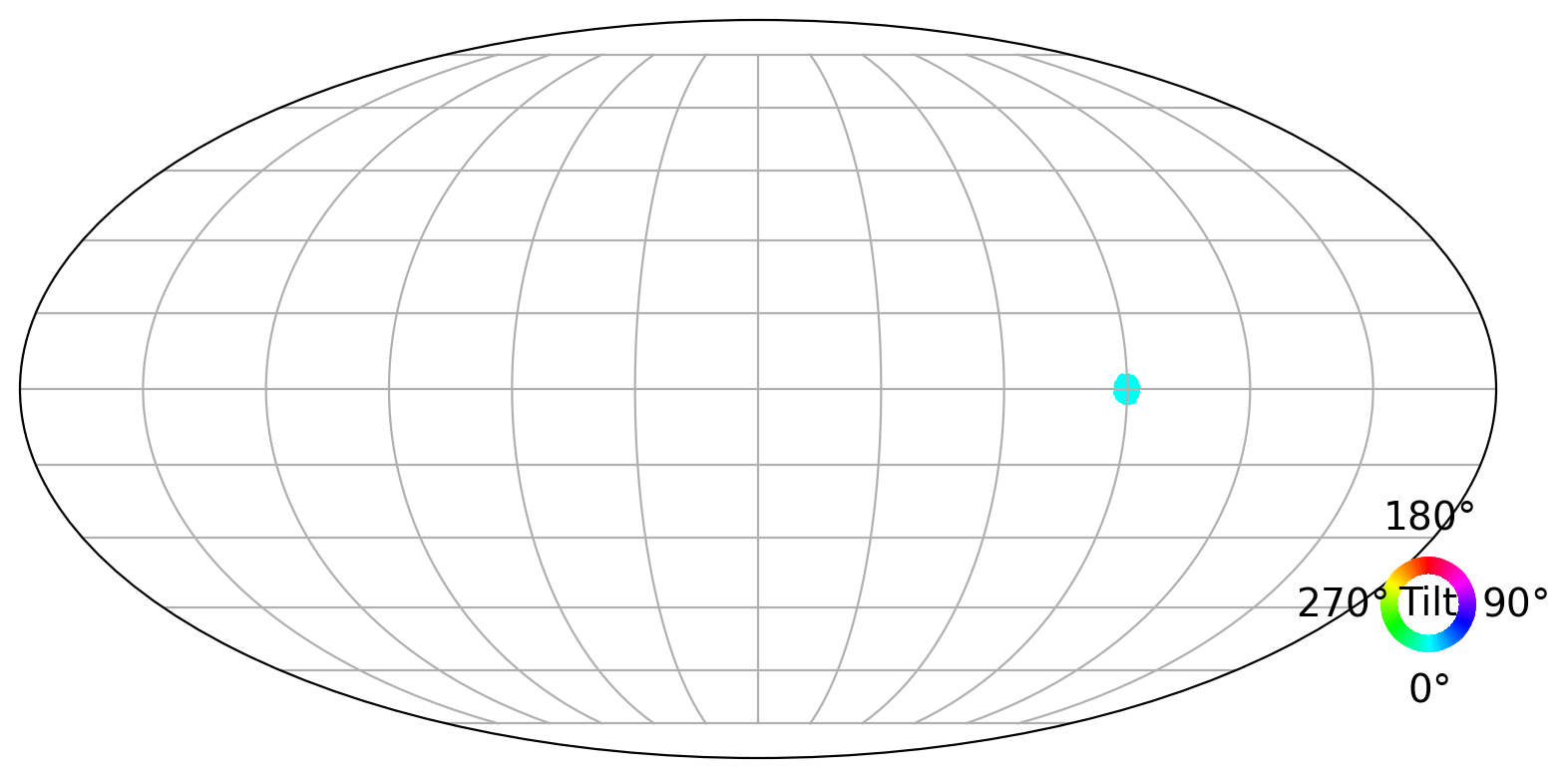} &
            \includegraphics[width=0.28\textwidth]{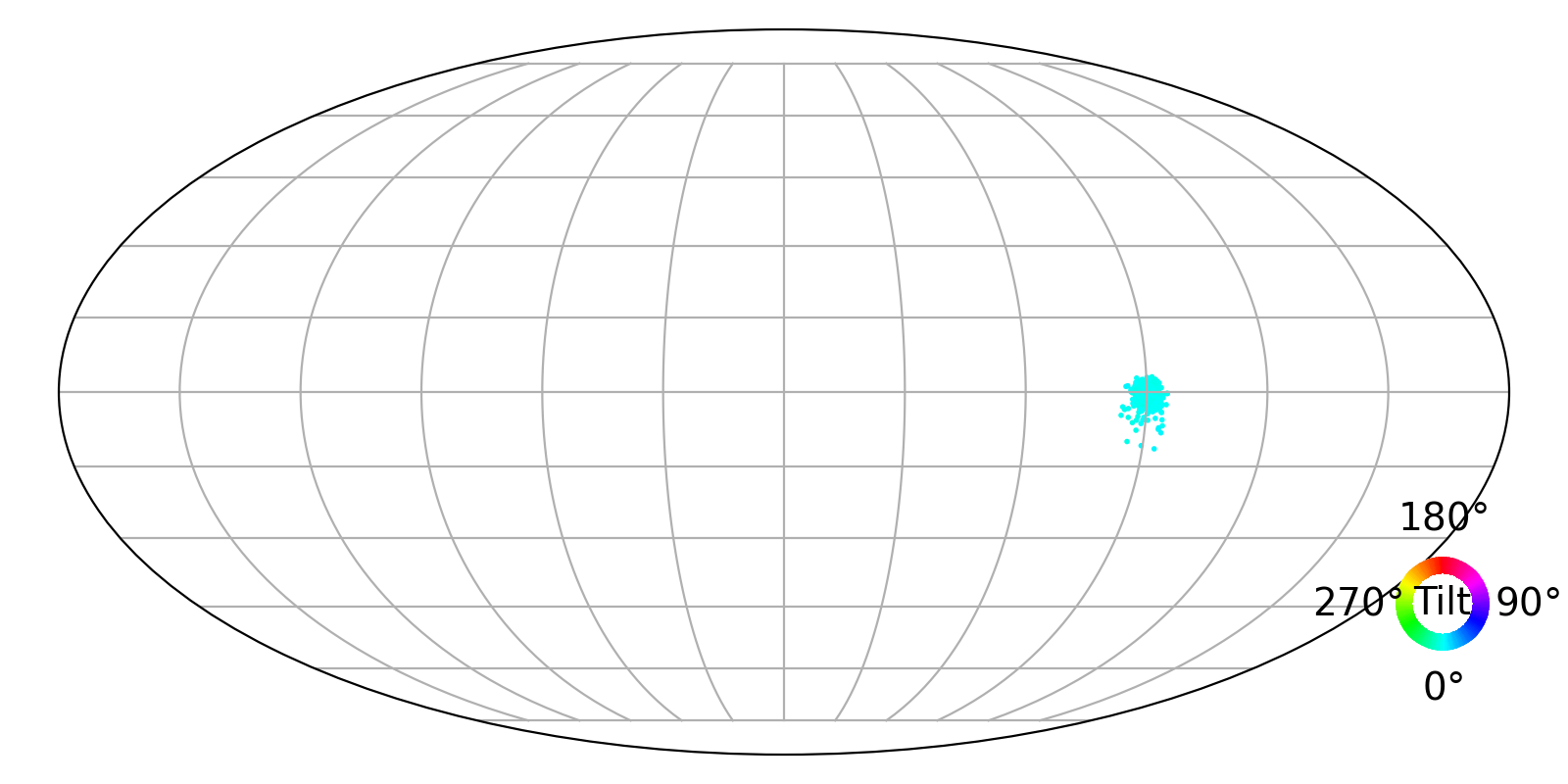} &
            \includegraphics[width=0.28\textwidth]{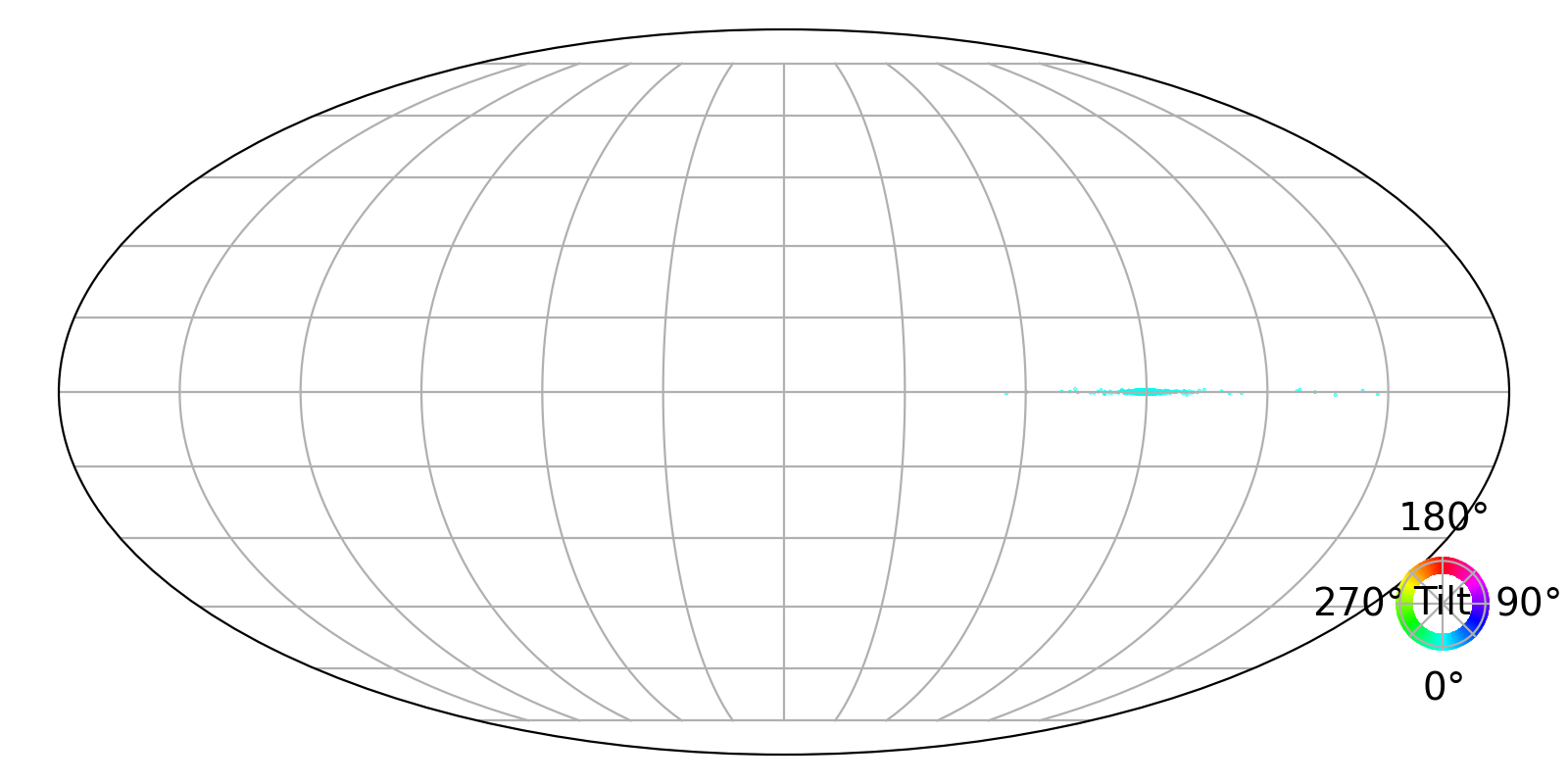} \\
            \includegraphics[width=0.28\textwidth]{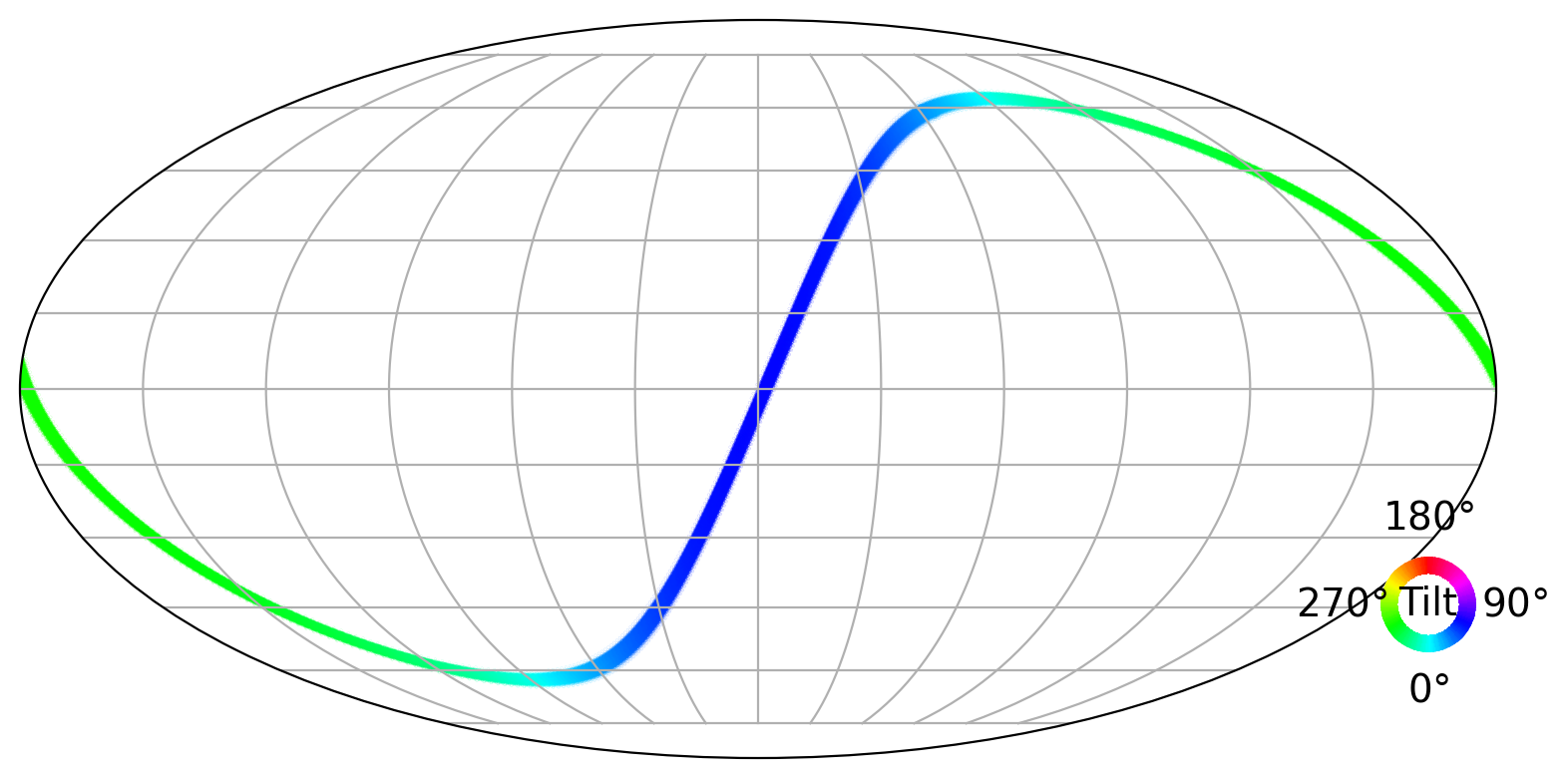} &
            \includegraphics[width=0.28\textwidth]{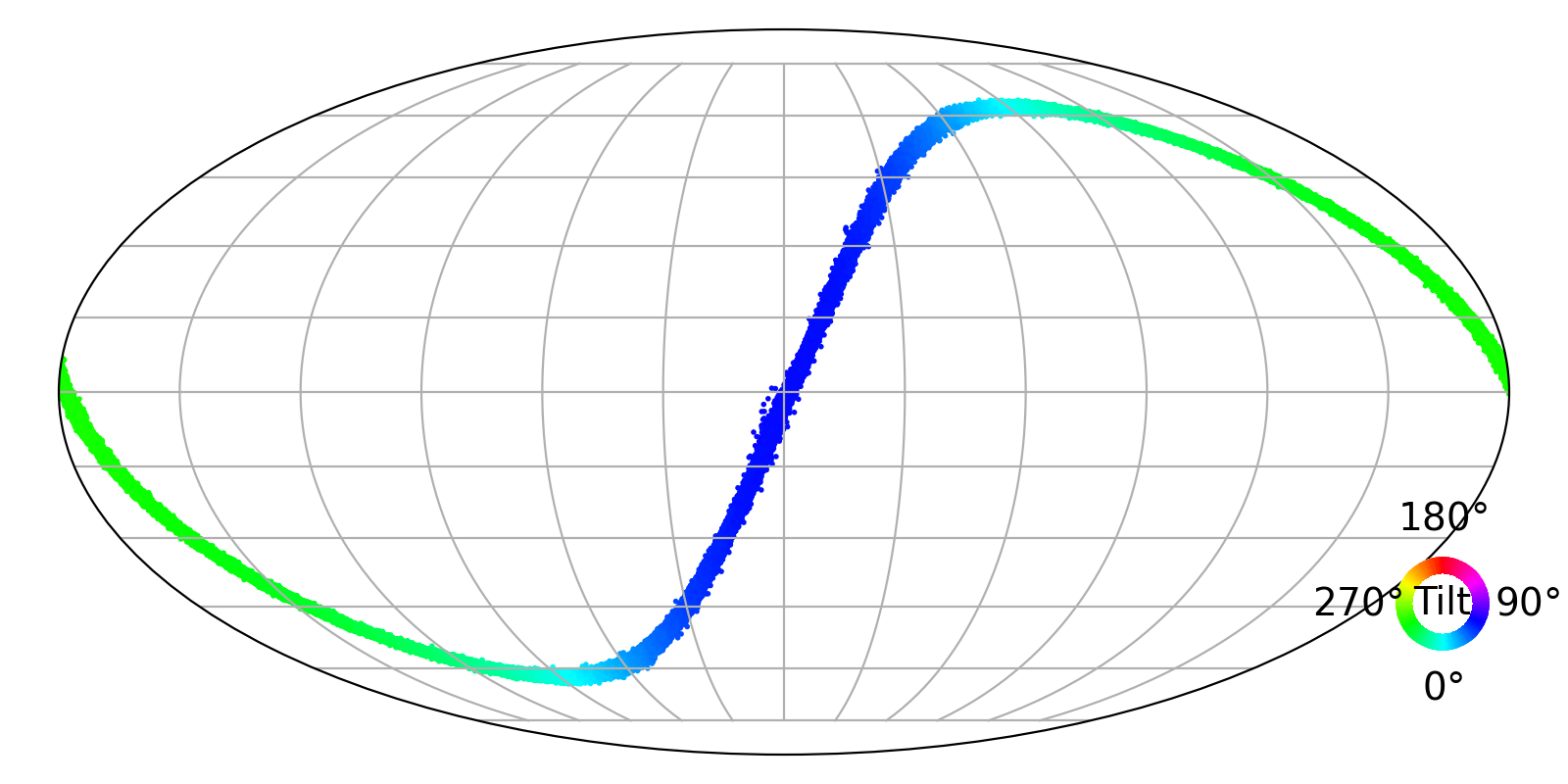} &
            \includegraphics[width=0.28\textwidth]{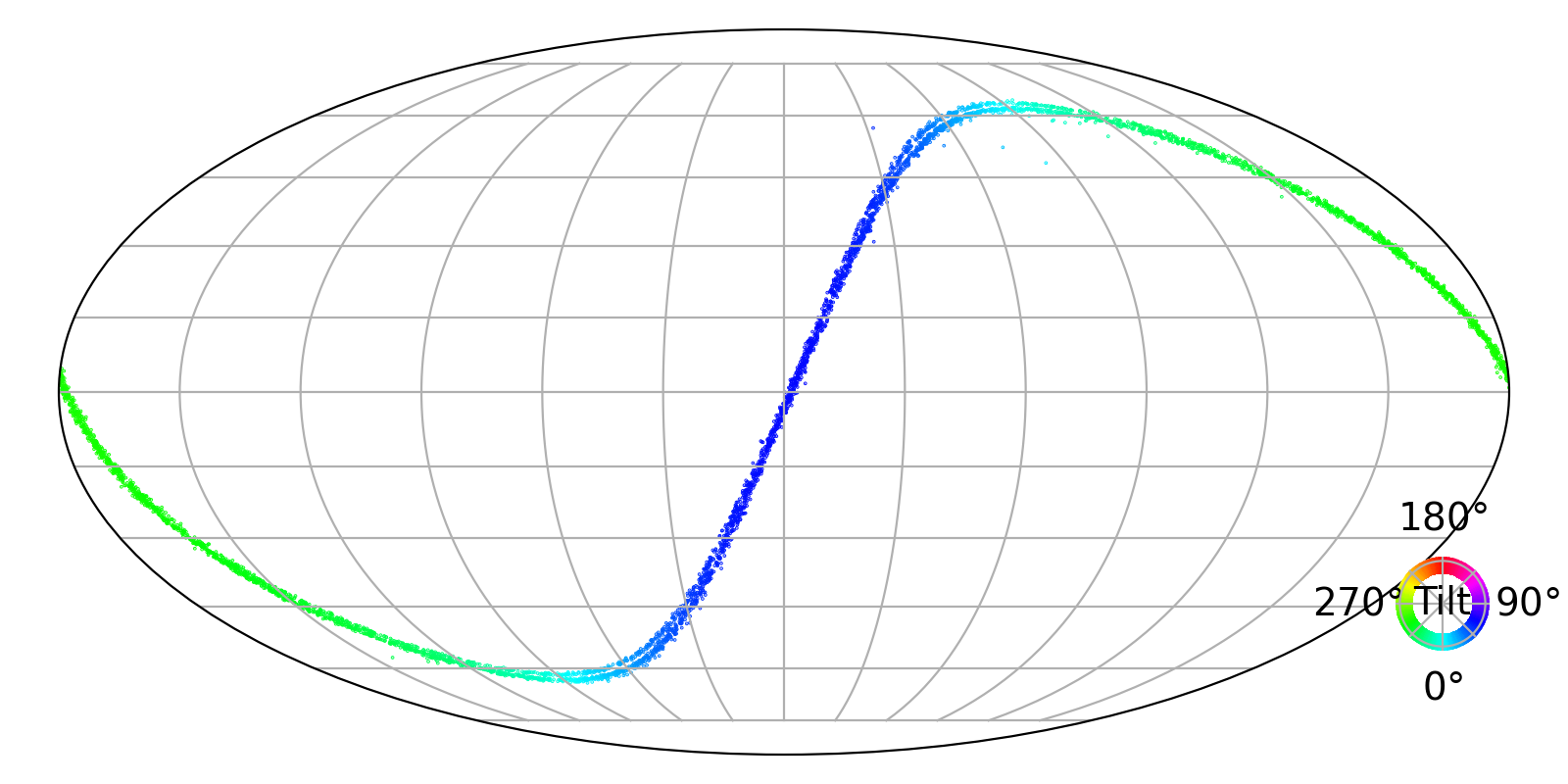} \\
            \includegraphics[width=0.28\textwidth]{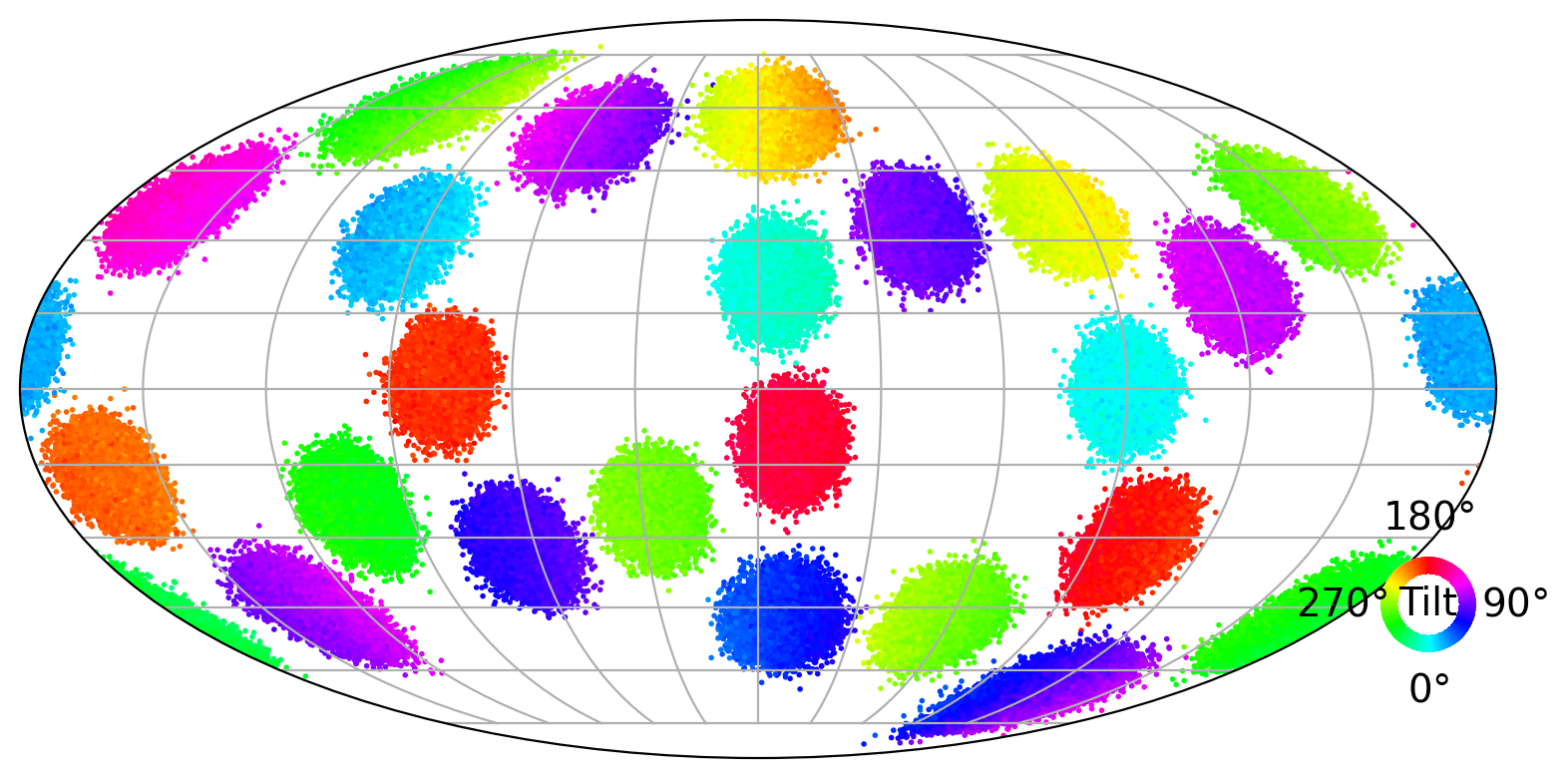} &
            \includegraphics[width=0.28\textwidth]{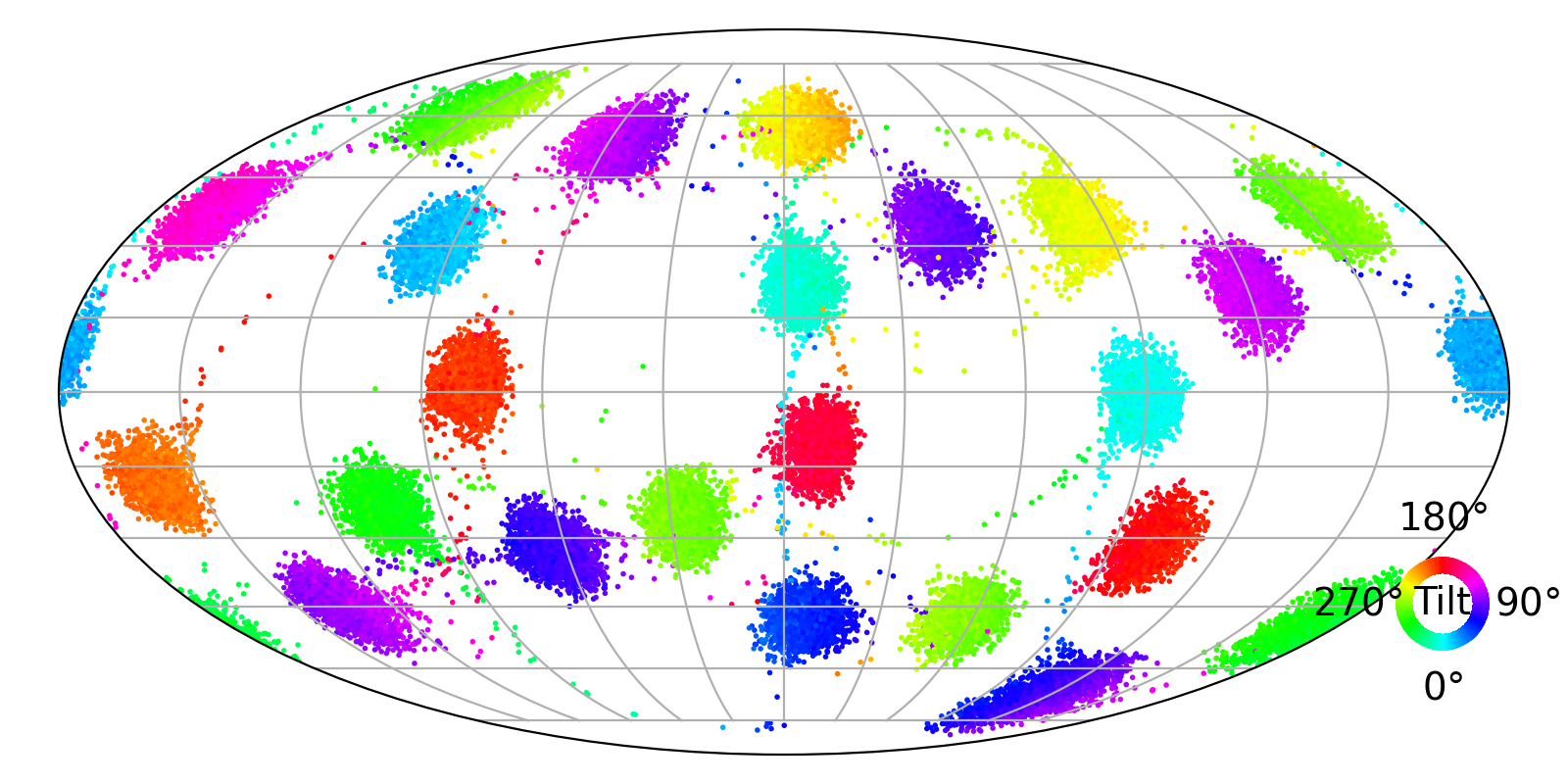} &
            \includegraphics[width=0.28\textwidth]{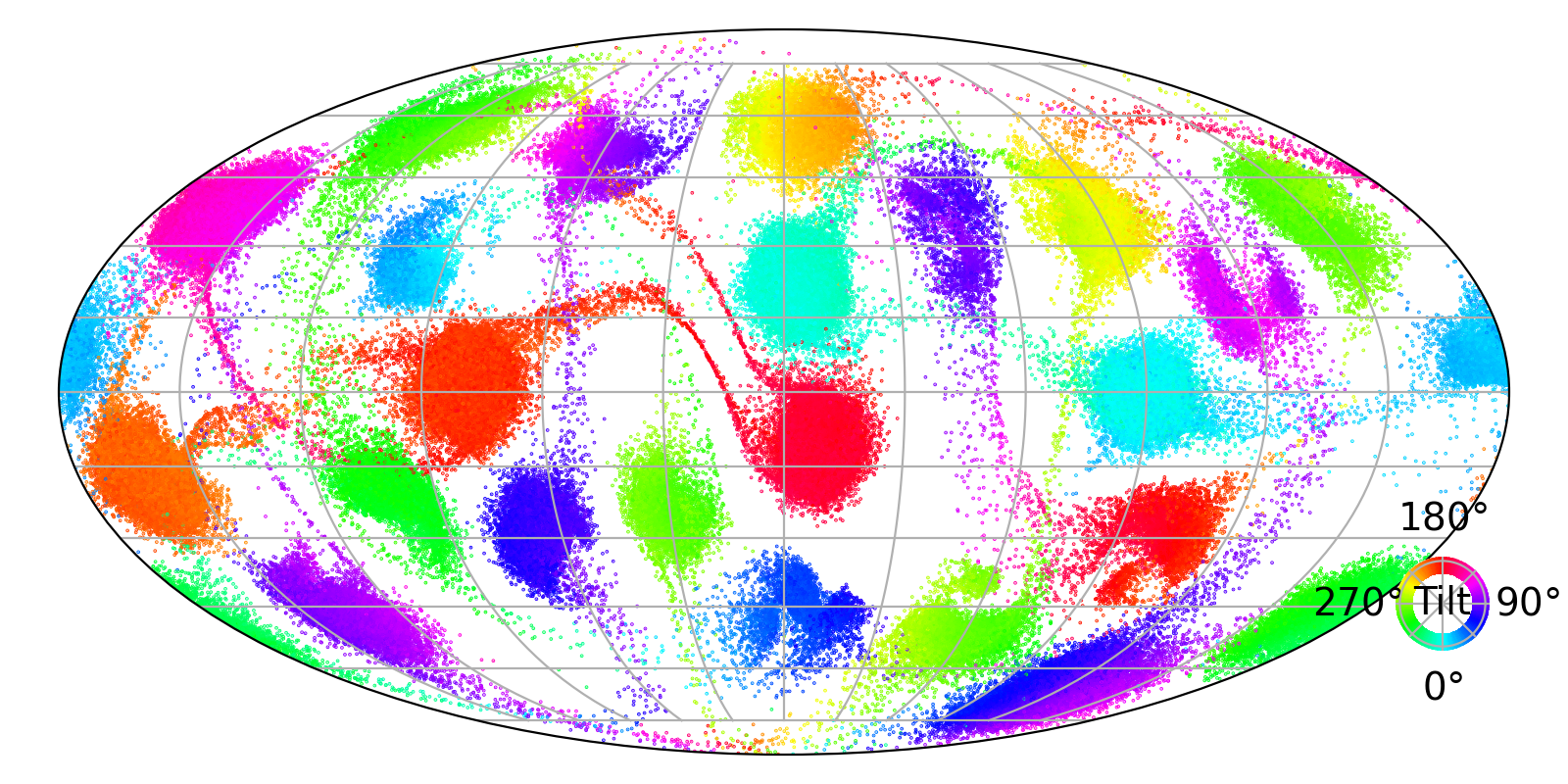} \\
            \\
        \end{tabular}
        \caption{
            \label{table:qualitative}
            Results for ``synthetic'' dataset.
            From top row to bottom row:
            Ground truth and results for subsets ``peak'', ``cone'', ``cube''.
            From left column to right column:
            Ground truth distribution, samples generated by the model proposed in~\cite{liu2023delving}, and samples generated by the proposed Euler angles-based flow model.
            The rotation matrix of~\cite{liu2023delving} tends to 
            model better disjoint ``islands'' of mass; 
            however, the proposed Euler angles model is better at estimating variance.
            (see also comment on~\cref{table:qualitative2}).
        }
    \end{minipage}
    \hfill
    \begin{minipage}{0.49\linewidth}
        \centering
        \begin{tabular}{ccc}
            \includegraphics[width=0.2\textwidth]{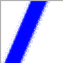} &
            \includegraphics[width=0.2\textwidth]{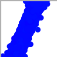} &
            \includegraphics[width=0.2\textwidth]{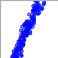} \\
            \includegraphics[width=0.2\textwidth]{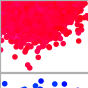} &
            \includegraphics[width=0.2\textwidth]{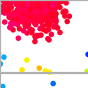} &
            \includegraphics[width=0.2\textwidth]{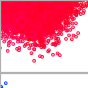} \\        
            \\
        \end{tabular}
        \caption{
            \label{table:qualitative2}
            Detail from Figure~\ref{table:qualitative}.
            From left column to right column:
            Ground truth distribution, samples generated by the model proposed in~\cite{liu2023delving}, and samples generated by the proposed Euler angles-based flow model.
            Note the difference in inductive biases between the two models:
            Liu \etal~\cite{liu2023delving} tends to misestimate variance by a considerable margin; the proposed Euler angles flow is closer to the correct value.
        }
    \end{minipage}
\end{figure}

\textbf{Synthetic dataset.}
\label{subsec:raw}
Each subset of this dataset is made up of a number of samples off the $\mathcal{SO}(3)$ manifold.
It was introduced in~\cite{liu2023delving}, and the quality of the fit to the data is measured in terms of log-likelihood, as shown in Table \ref{table:synthetic}.
We use $24$ Euler angle-based M\"obius flow layers,
each of which uses $K=64$ M\"obius kernels.
A batch size of $1024$ and learning rate equal to $10^{-4}$ is used.
Models are trained for $50k$ iterations.
Samples from learned distributions can be examined in~\cref{table:qualitative} and~\ref{table:qualitative2}.
We use the visualisation tool of \cite{murphy2021implicit}.
The two models -- the rotation matrix-based~\cite{liu2023delving} and the proposed Euler angles flow model -- exhibit distinctly different inductive biases.
We can discern two kinds of differences. First, the rotation matrix model handles cases where the PDF is made up of disjoint islands of mass.
Second, the Euler angles model tends to estimate variance more faithfully.
Overall, these differences stack up to a slight advantage for the Euler angles implementation, as per the numerical results of Table \ref{table:synthetic}.

\textbf{Gimbal dataset.}
We introduce an additional dataset for unsupervised density estimation that we call \textit{Gimbal}, in order to gauge the efficiency of the proposed Euler flows exactly with areas of $\SO3$ that are problematic for our parameterisation.
Regardless of the choice of axes in an Euler angles parameterisation,
there will always exist parts of $\SO3$ where gimbal lock takes place
(cf. discussion in Section \ref{sec:proposed}).
Under the convention we use,
this happens when $\phi = \frac{\pi}{2}$ or $\phi = -\frac{\pi}{2}$.
We produce random elements of $\SO3$ by sampling as follows:
    $\omega \sim \mathcal{N}(0,\sigma^2)$,
    $\phi \sim \frac{1}{2}\mathcal{N}(\frac{\pi}{2}, \sigma^2_\phi\sigma^2) + \frac{1}{2}\mathcal{N}(-\frac{\pi}{2}, \sigma^2_\phi\sigma^2)$,
    $\kappa \sim \mathcal{N}(0,\sigma^2)$,
where we set $\sigma_\phi^2 = 0.1$.
We set $\sigma^2 = \{1, 0.1\}$.
Following the layout of the ``synthetic'' dataset discussed in the previous subsection, we split the sampled rotations into a training and test set that are made up of $60k$ and $12k$ iterations, respectively.
We have compared the proposed model versus the model of Liu \etal~\cite{liu2023delving}, using the setups described in subsection \ref{subsec:raw}.
Numerical results are reported in Table \ref{table:gimbal}.
We can deduce that the more data is concentrated around areas of singularity,
the harder it is for the Euler angles model to achieve a good fit.
\begin{table}[h]
\centering
\caption{
Results for ``gimbal'' dataset.
Testing of the proposed Euler angles flow model in areas of singularity.
Log-likelihood is used to evaluate the quality of fit.
\label{table:gimbal}}
\begin{tabular}{l c c}
\\ \toprule
log likelihood $\uparrow$ & {$\sigma^2$=1} & {$\sigma^2$=0.1}\\
\midrule
M\"obius + Affine Flow~\cite{liu2023delving} & \textbf{4.50} & \textbf{13.31} \\
Euler angles flow & 2.74 & 9.46 \\
\bottomrule
\end{tabular}
\end{table}
It is worth noting that here we see a different picture than the one of the results
of the previous synthetic experiment (Table~\ref{table:synthetic}),
as there is a stark difference in log-likelihood, but now
Euler angles are the model that fares less competitively.
This is perhaps not surprising, as the data of this set
cover deliberately the areas of $\SO3$ related to Euler angles singularity.

\textbf{SYMSOL I dataset.}
We perform experiments on SYMSOL I~\cite{murphy2021implicit},
which comprise sets of objects with pose ground truths that are inherently multimodal due to object symmetries.
SYMSOL I contains high-order symmetry shape images such as tetrahedron, cube, cone, and cylinder.
We train our model for $900k$ iterations and otherwise follow the training regime for the M\"obius and Quaternion affine-based models of Liu \etal~\cite{liu2023delving}.
For our implementation, we use the same flow step sequence where we have disabled Quaternion affine flow layers.
Otherwise, the default parameters for running the code were used.
The results 
are presented in Table~\ref{tab:rotation_regression}.
Our method achieves the best log-likelihood performance for the cone and cylinder shapes and the second-best for the cube shape, resulting in the second-best average log-likelihood performance. 
As a comment on the results, it seems that there is a negative trend
on objects which have ground truth pdfs with numerous disjoint islands of mass.
Such are objects like the icosahedron or the tetrahedron.
This is perhaps related to a similar effect in the ``Synthetic'' dataset (cf. Figure~\ref{table:qualitative} \& discussion in Section \ref{sec:conclusion}).

\begin{table}[h]
    \centering
    \begin{minipage}{0.51\linewidth}
        \centering
        \caption{
            Numerical results on SYMSOL I.  
            Exponents $M$ and $M+A$ correspond to using \cite{liu2023delving} with only M\"obius layers
            or the full model.
            Best and second-best performance are highlighted with boldface and underline, respectively.  
        }
        \label{tab:rotation_regression}
        \setlength{\tabcolsep}{2pt}
        \begin{tabular}{lcccccc}
        \\ \toprule
        & \multicolumn{6}{c}{log likelihood $\uparrow$} \\
        \cmidrule(lr){2-7}
        & avg. & cone & cube & cyl. & ico. & tet. \\
        \midrule
        \cite{deng2021bingham} & 0.81 & 2.45 & -2.15 & 1.34 & -0.16 & 2.56 \\
        \cite{gilitschenski2019bingham} & 1.86 & 6.13 & 0.00 & 3.17 & 0.00 & 0.00 \\
        \cite{prokudin2018vmf} & 0.42 & -1.05 & 1.79 & 1.01 & -0.10 & 0.43 \\
        \cite{murphy2021implicit} & 6.39 & 6.74 & 7.10 & 6.55 & 3.57 & 7.99 \\
        \cite{liu2023delving}$^M$ & 9.41 & \underline{10.52} & 9.68 & \underline{10.00} & \underline{5.35} & \underline{11.51} \\
        \cite{liu2023delving}$^{M+A}$ & \textbf{10.38} & 10.05 & \textbf{11.64} & 9.54 & \textbf{8.26} & \textbf{12.43} \\
        \hline
        Ours & \underline{9.66} & \textbf{12.36} & \underline{10.20} & \textbf{11.64} & 4.15 & 9.97 \\
        \bottomrule
        \end{tabular}
    \end{minipage}
    \hfill
    \begin{minipage}{0.46\linewidth}
        \centering
        \caption{
            Numerical results of ModelNet10-SO3. 
            Indices $U$ and $F$ denote the base distribution used for the flow in each case. 
            Best and second-best performance is highlighted, with boldface and underline, respectively.
        }
        \label{table:modelnet}
        \setlength{\tabcolsep}{2pt}
        \begin{tabular}{lccc}
            \\ \toprule
            {} & Acc@\(\,15^{\circ}\)\(\uparrow\) & Acc@\(\,30^{\circ}\)\(\uparrow\) & Med.\ (\(^{\circ}\))\(\downarrow\) \\ 
            \midrule
            \cite{deng2021bingham} & 0.562 & 0.694 & 32.6 \\
            \cite{prokudin2018vmf} & 0.456 & 0.528 & 49.3 \\
            \cite{mohlin2020mfisher} & 0.693 & 0.757 & 17.1 \\
            \cite{murphy2021implicit} & 0.719 & 0.735 & 21.5 \\
            \cite{liu2023delving}$_U$ & \textbf{0.760} & \textbf{0.774} & 14.6 \\
            \cite{liu2023delving}$_F$ & \underline{0.744} & \underline{0.768} & \textbf{12.2} \\
            \midrule
            Ours$_{U}$\!\!\!\!\!\! & 0.742 & 0.753 & 19.5\\
            Ours$_{F}$\!\!\!\!\!\! & 0.736 & 0.754 & \underline{12.7}\\
            \bottomrule
        \end{tabular}
    \end{minipage}
\end{table}

\textbf{ModelNet10-SO3 dataset.}
We further conduct experiments on the ModelNet10-SO3 dataset~\cite{liao2019spherical} that is widely used for regressing rotations from single images.
The dataset is comprised of the uniformly rotated renderings of CAD models from the ModelNet10 dataset~\cite{wu20153d}.
We present the performance using both the uniform and pre-trained Fisher distributions in terms of Acc@\(\,15^{\circ}\)\,, Acc@\(\,30^{\circ}\)\,, and Median Error in Table~\ref{table:modelnet}.   
As we can see, our proposed approach achieves competitive accuracies, close to the highest performing methods, while we achieve the second-best median error after~\cite{liu2023delving} with the corresponding Fisher distribution.

\subsection{Training speed comparison}
\textbf{Training speed comparison.}
We report a training speed comparison over the proposed Euler flows alternative,
versus two of the versions presented in Liu \etal~\cite{liu2023delving}.
Speed tests were run on an RTX 2080Ti GPU.
Tests on unsupervised density estimation (synthetic) use default settings,
and conditional estimation uses a batch size of 2.
Results are shown in Table \ref{table:speed}.
The speed of the proposed model can be attributed to the simpler architecture of the flow,
owing to the minimal character of the Euler angles representation.




\begin{table}[h]
\centering
\caption{
Training speed comparison:
Reported figures are milliseconds per training iteration, on average.
Exponents $M$ and $M+A$ correspond to using \cite{liu2023delving} with only M\"obius layers or the full model.
See text for details.
\label{table:speed}}
\scalebox{0.96}{
\begin{tabular}{l c c}
\\ \toprule
Time per iteration (ms) $\downarrow$ & {Unconditional} & {Conditional}\\
\midrule
\cite{liu2023delving}$^{M}$ & 483 & 384 \\
\cite{liu2023delving}$^{M+A}$ & 966 & 734 \\
Ours & \textbf{387} & \textbf{335} \\
\bottomrule
\end{tabular}}
\end{table}

\vspace{-1cm}
\section{Conclusion and Future Work}
\label{sec:conclusion}

As a one-sentence answer to the titular question of this paper,
``Are Euler angles a useful parameterisation 
of rotation
for pose estimation with Normalizing Flows?'' 
\emph{Yes}, 
despite their shortcomings, they can be used to construct a useful model.
Our experiments corroborate this point of view, and numerical results are routinely in the ballpark of the state of the art.
However, our analysis has shown that there are several important nuances that must be taken into account if we are to decide whether we should use this model in a practical scenario versus a model that does not use this parameterisation.
There do exist caveats, some of which were not completely surprising.
In particular, the proposed model does not fare well in areas related to gimbal lock singularity.
In a practical scenario, this, of course, must be taken into account; as with the use of Euler angles, 
the key factor here is whether or not we have prior knowledge that our angles are 
related to areas of singularity (cf. the case of aerophotogrammetry).
A point that we believe is worth mentioning is related to the reasons for which we have slightly better results in many cases. 
This appears to be related to two factors: a) a different inductive bias than the one related to the rotation matrix model;
b) a learning-wise ``easier'' objective.
The Euler angles model seems to have some difficulty modeling those cases where there are disjoint masses in the ground truth.
This is a well-known issue with Normalizing Flows in general, and there is current research that may lead to considerable improvements in this respect \cite{bevins2024piecewise}.
Another point that is noteworthy is that Euler angles lend to a much less costly model, as it can be trained faster than both alternatives of the rotation matrix that we compared against.
Further, we have seen that the Euler angles model outperforms routinely other non-NF-based models for rotation estimation.
Finally, let us note that the current implementation using Euler angles is not the only way that one can construct a flow with them, as there are several hyperparameters in place.
First and foremost is the M\"obius transform itself, or the way we implement coupling (one dimension for the transformer versus two for the conditioner), putting in place phase-translating flows to improve expressivity, and so on.
Other considerations could include integration with more capable transformer backbones or fusing with other types of probabilistic models \cite{sfikas2011majorization,zhai2024normalizing}.
This could be the subject of future work.



\pagebreak

\section*{Acknowledgements}

The publication/registration fees were partially covered by the University of West Attica.

{
    \bibliography{refs}
}

\end{document}